\documentclass[lettersize,journal]{IEEEtran}

\usepackage{graphicx}
\usepackage{amsmath}
\usepackage{amssymb}
\usepackage{booktabs}
\usepackage{adjustbox}
\usepackage[dvipsnames,table]{xcolor}
\usepackage[ruled,vlined]{algorithm2e}
\usepackage[pagebackref,breaklinks,colorlinks]{hyperref}

\usepackage[capitalize]{cleveref}
\crefname{section}{Sec.}{Secs.}
\Crefname{section}{Section}{Sections}
\Crefname{table}{Table}{Tables}
\crefname{table}{Tab.}{Tabs.}

\begin{document}
\title{Language-Driven Active Learning for Diverse Open-Set 3D Object Detection}
\author{Ross~Greer*~\IEEEmembership{Member,~IEEE},
        Bj\o rk Antoniussen*~\IEEEmembership{Member,~IEEE,}, 
        Andreas M\o gelmose~\IEEEmembership{Member,~IEEE,},
        Mohan~M.~Trivedi~\IEEEmembership{Life Fellow,~IEEE}%\vspace{-1cm}% <-this % stops a space
\thanks{*Authors R. Greer and B. Antoniussen contributed equally. R. Greer, B. Antoniussen, and M. Trivedi are with the Laboratory for Intelligent and Safe Automobiles at the University of California San Diego. B. Antoniussen and A. Møgelmose are with the Visual Analysis and Perception Lab at Aalborg Universitet. Corresponding Author E-mail: regreer@ucsd.edu.}% <-this % stops a space
% %\thanks{B. Antoniussen, M. Andersen, and A. Møgelmose are with the Faculty of Electronic Systems, Aalborg University.}
}
% \author{Ross Greer \thanks{Authors contributed equally. R. Greer, B. Antoniussen, and M. Trivedi are with the Laboratory for Intelligent \& Safe Automobiles (LISA) at University of California San Diego. A. Møgelmose is with Aalborg University.} \\
% {\tt\small regreer@ucsd.edu}
% \and
% Bjørk Antoniussen *\\
% {\tt\small banton19@student.aau.dk}
% \and
% Andreas Møgelmose \\
% {\tt\small anmo@create.aau.dk}
% \and
% Mohan Trivedi\\
% {\tt\small mtrivedi@ucsd.edu}
% }

%\vspace{-9cm}

\maketitle
\markboth{Journal of \LaTeX\ Class Files,~Vol.~14, No.~8, August~2015}%
{Shell \MakeLowercase{\textit{et al.}}: Bare Demo of IEEEtran.cls for IEEE Journals}



%\maketitle

\begin{abstract}
Object detection is crucial for ensuring safe autonomous driving. However, data-driven approaches face challenges when encountering minority or novel objects in the 3D driving scene. In this paper, we propose VisLED, a language-driven active learning framework for diverse open-set 3D Object Detection. Our method leverages active learning techniques to query diverse and informative data samples from an unlabeled pool, enhancing the model's ability to detect underrepresented or novel objects. Specifically, we introduce the Vision-Language Embedding Diversity Querying (VisLED-Querying) algorithm, which operates in both open-world exploring and closed-world mining settings. In open-world exploring, VisLED-Querying selects data points most novel relative to existing data, while in closed-world mining, it mines novel instances of known classes. We evaluate our approach on the nuScenes dataset and demonstrate its efficiency compared to random sampling and entropy-querying methods. Our results show that VisLED-Querying consistently outperforms random sampling and offers competitive performance compared to entropy-querying despite the latter's model-optimality, highlighting the potential of VisLED for improving object detection in autonomous driving scenarios. We make our code publicly available at \url{https://github.com/Bjork-crypto/VisLED-Querying}
\end{abstract}

\section{Introduction}
\begin{figure}
    \centering

    \includegraphics[width=.24\textwidth]{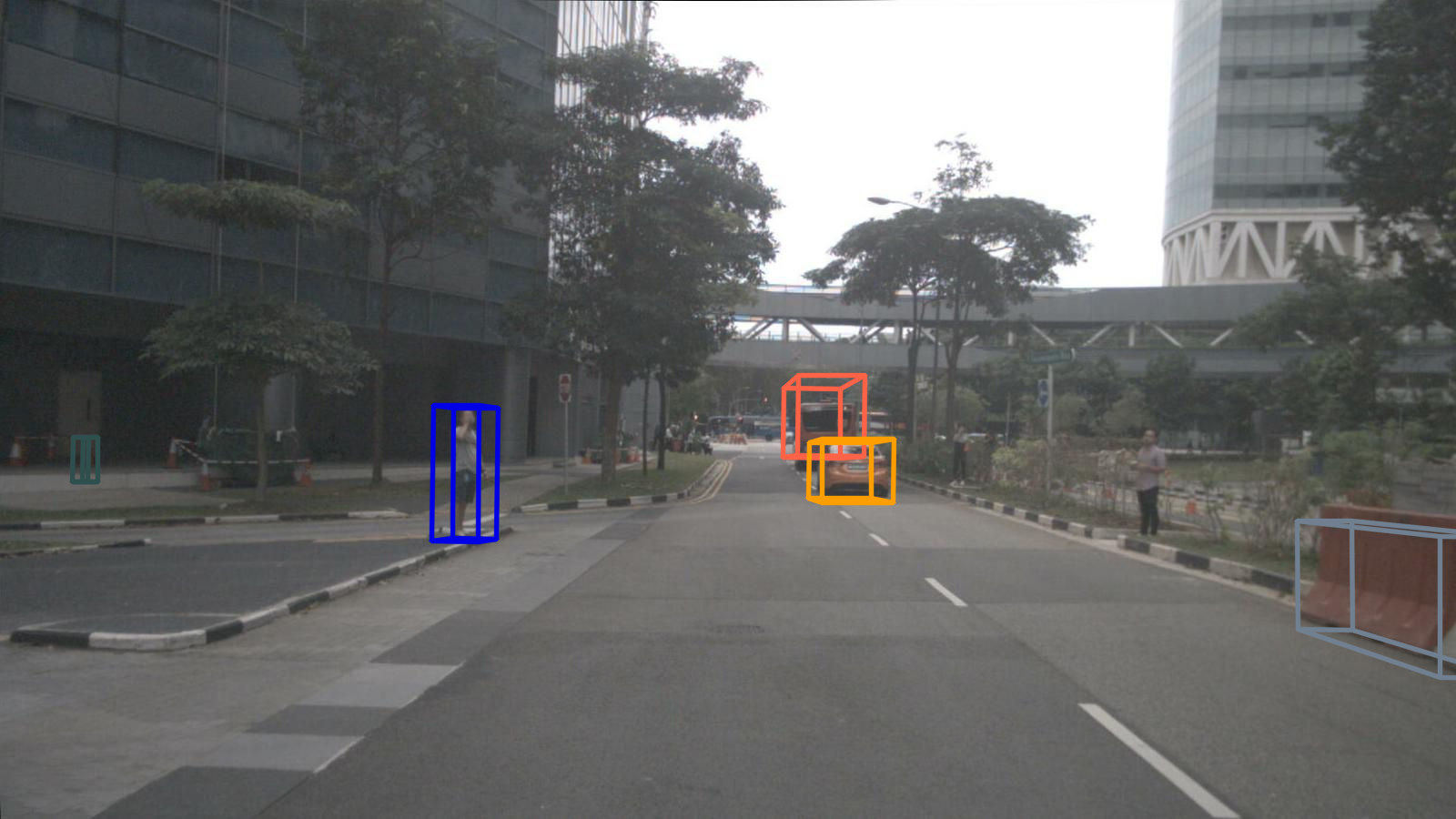}
    \includegraphics[width=.24\textwidth]{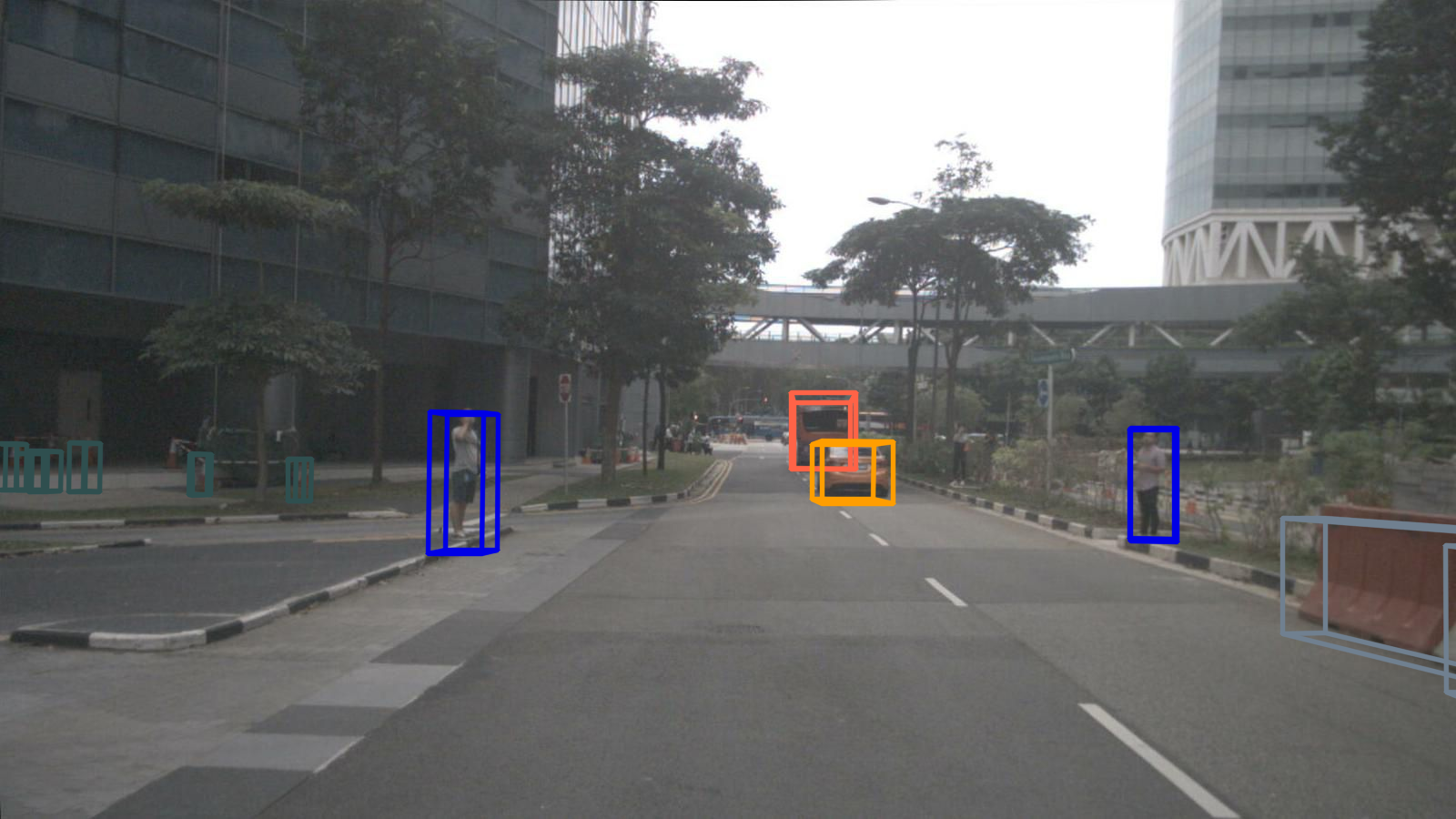}
    \includegraphics[width=.24\textwidth]{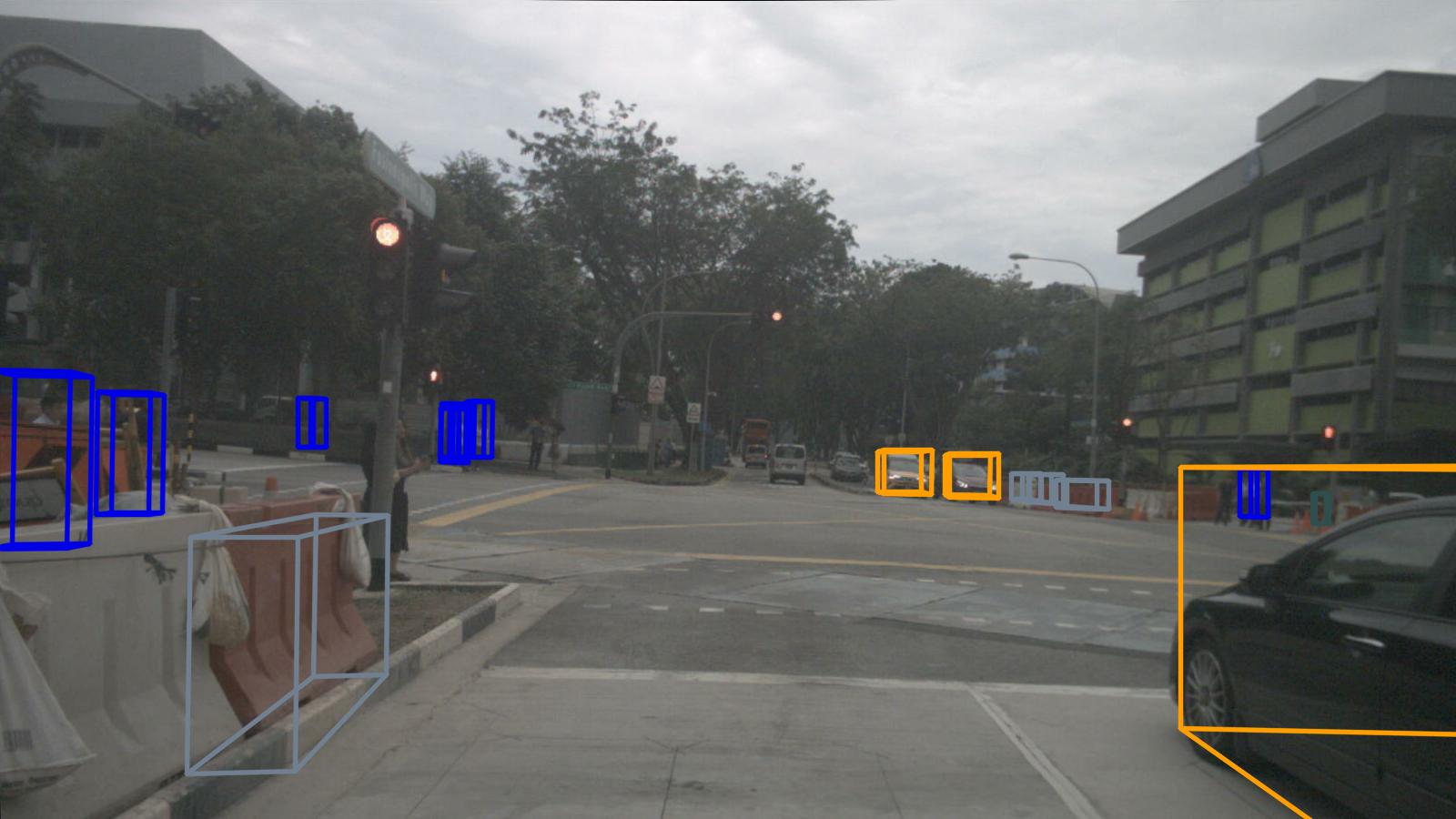}
    \includegraphics[width=.24\textwidth]{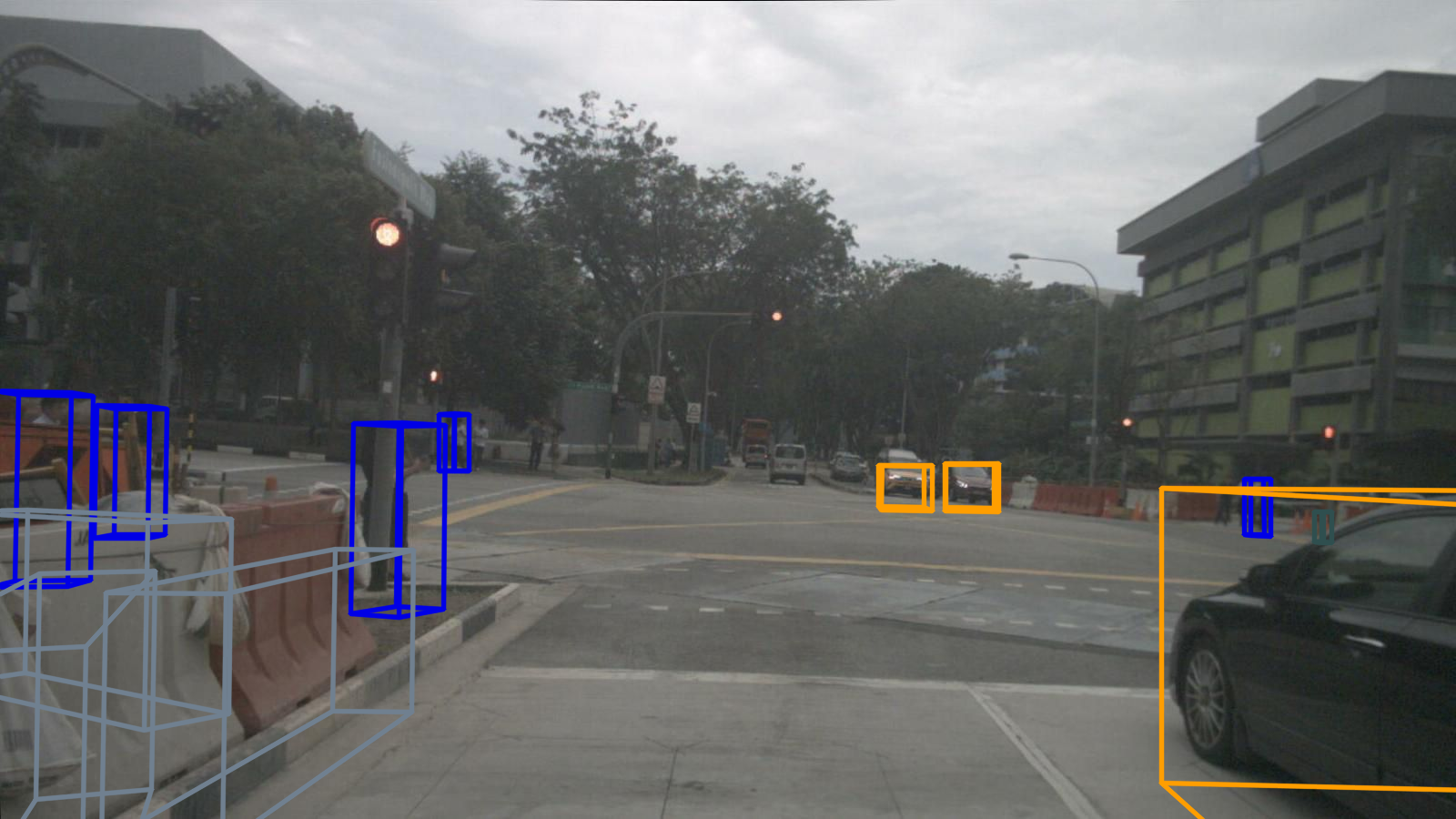}
    \includegraphics[width=.24\textwidth]{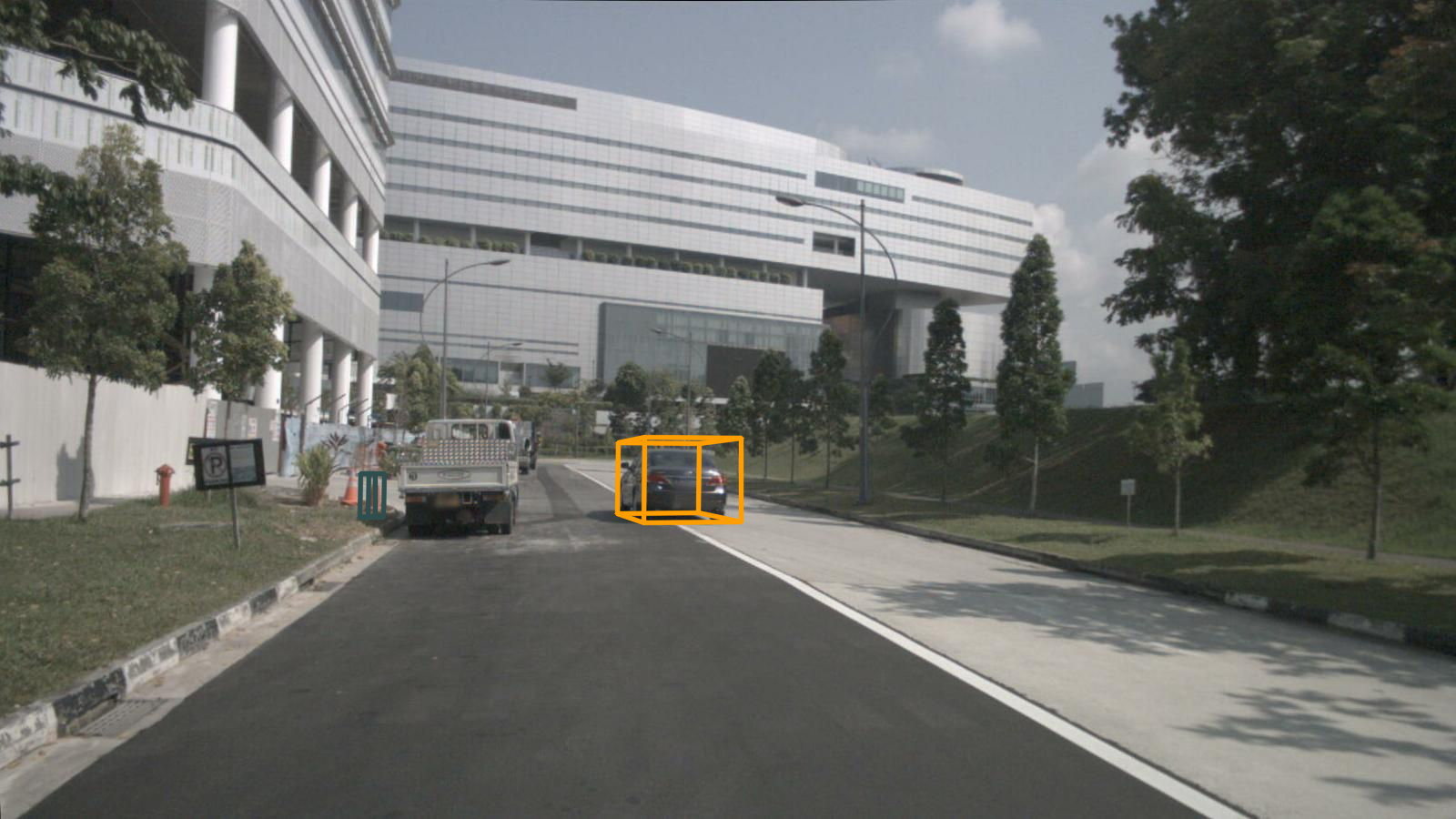}
    \includegraphics[width=.24\textwidth]{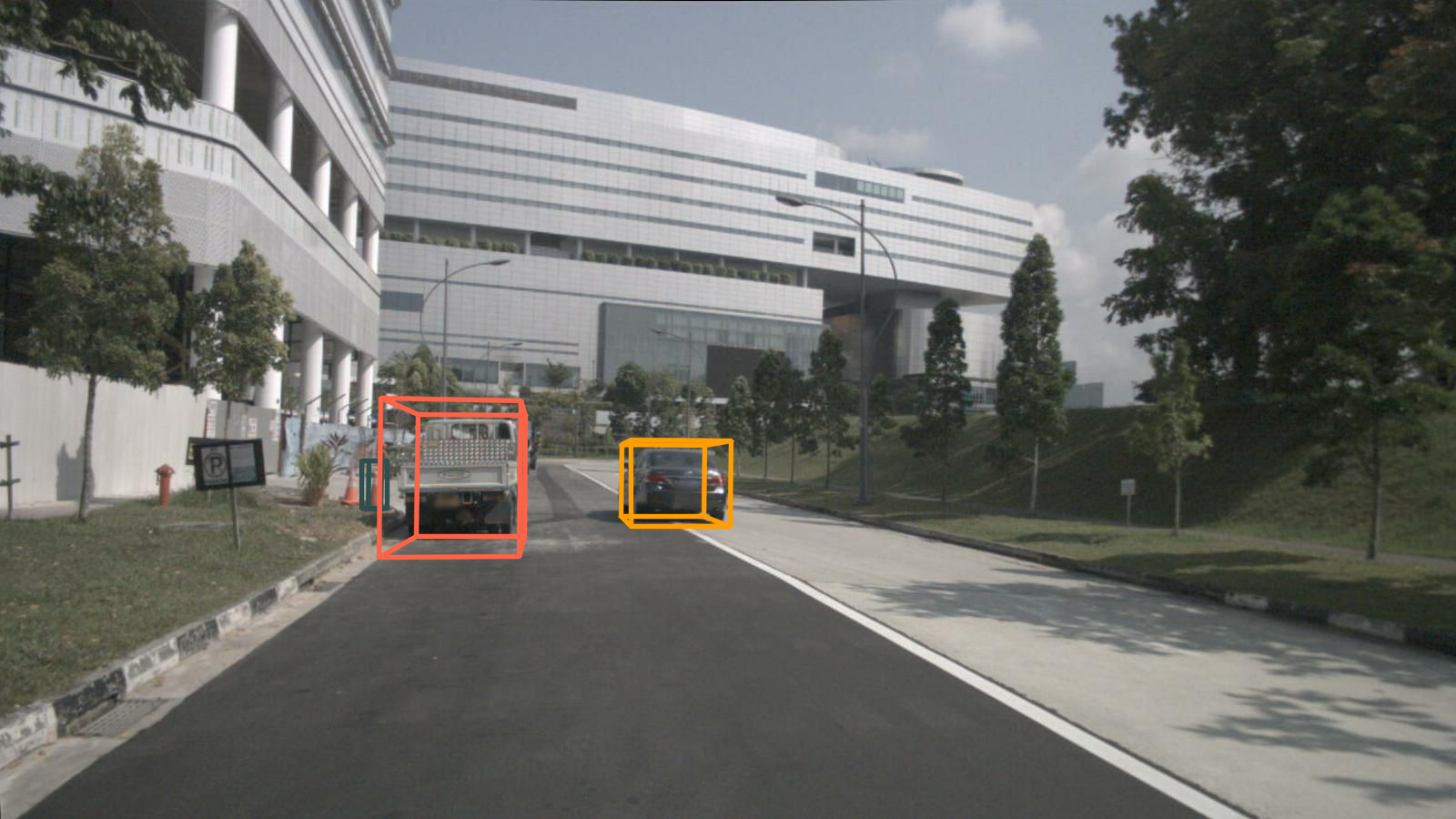}
    \includegraphics[width=.24\textwidth]{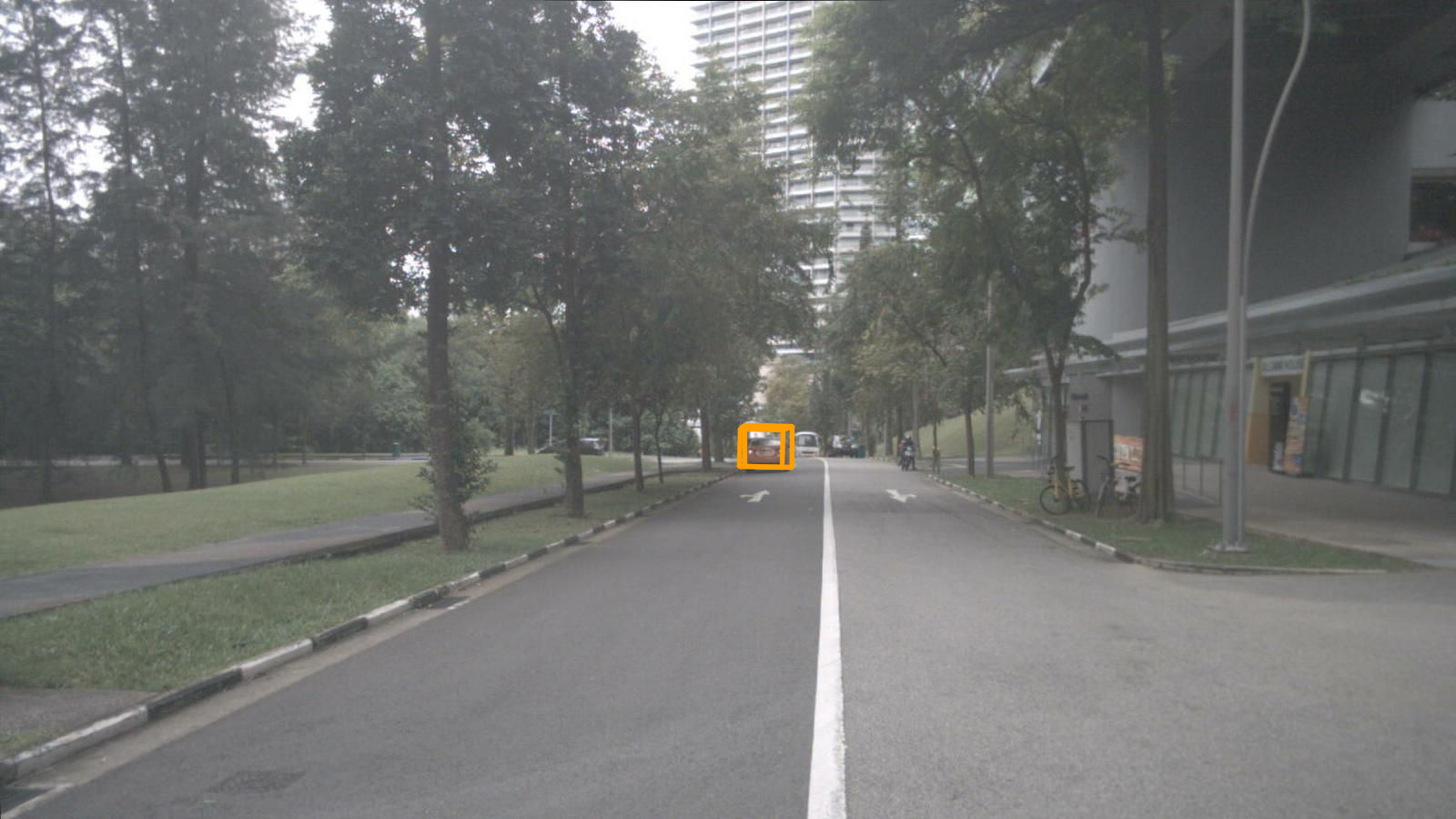}
    \includegraphics[width=.24\textwidth]{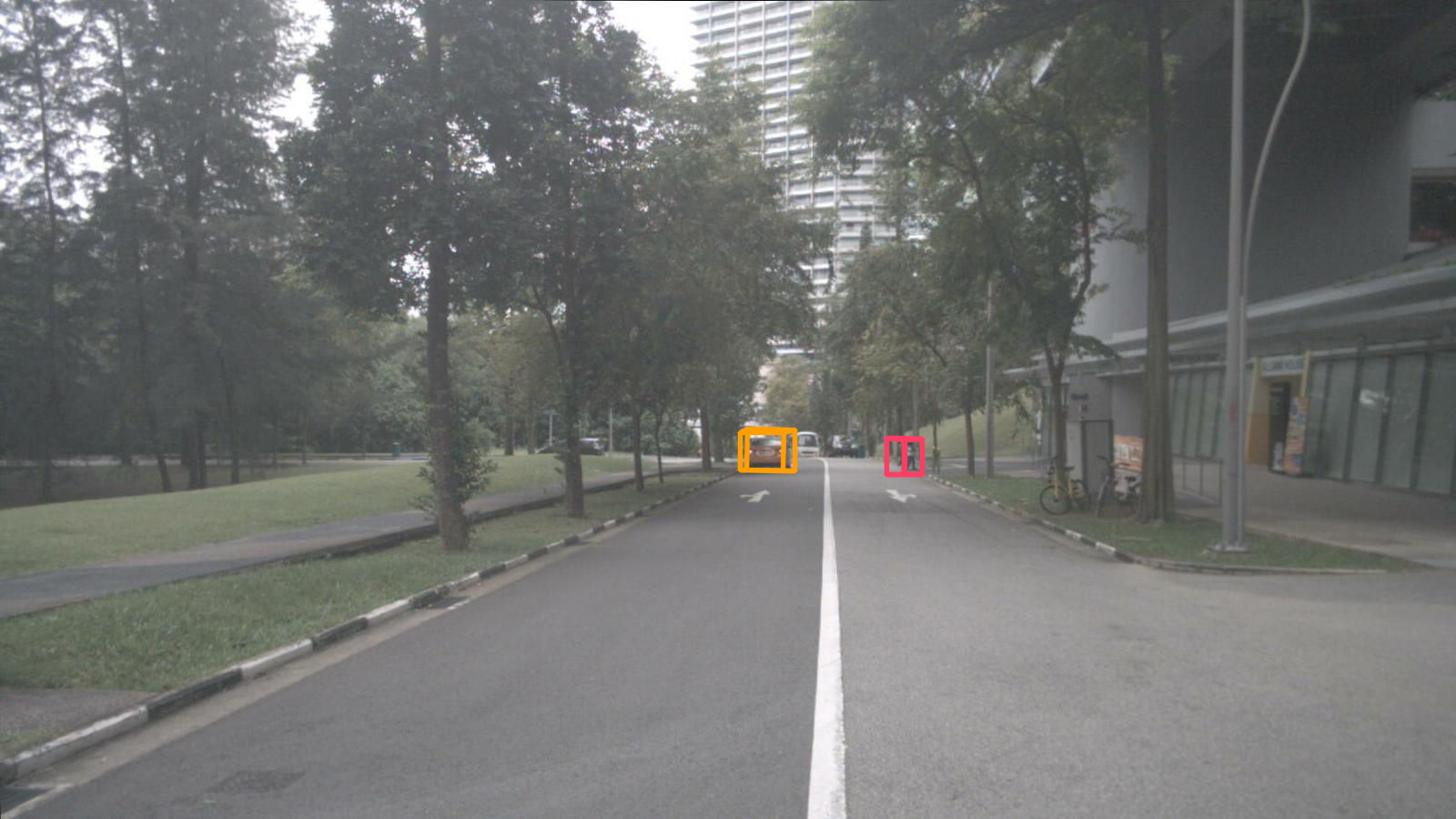}
    \caption{Choosing the most informative data can impact object detection model performance. Images in the left column are the results of a model trained on 50\% of nuScenes data, selected at random. Images in the right column are the results on the same images of a model trained on 50\% of nuScenes data, but selected using our VisLED active learning query strategy. In the top two rows, we see cases where challenging pedestrians are missed on the left image (preparing to cross on the right side of the road, and standing behind the crossing pole, respectively), but correctly detected on the right. Similarly, in the bottom two rows, the under-represented classes of motorcycle and truck are more readily detected using our active learning strategy.}
    \label{fig:beginning_examples}
\end{figure}

Object detection is critical for safe autonomous driving. Data-driven approaches currently provide the best performance in detecting and localizing objects in the 3D driving scene. Detection models perform best on objects which are most represented in driving datasets. This creates challenges when some objects are less represented (minority classes), or unrepresented within the annotation scheme (``novel" objects \cite{chen2020task}, relevant for ``open-set" learning \cite{scheirer2012toward}), and becomes especially important when minority objects are most salient to driving decisions \cite{greer2022salience, ohn2016makes, greer2023salient, greer2023robust}. Further, from a pragmatic standpoint, the collection, curation, and annotation of such datasets can be extremely expensive \cite{behl2020label, kulkarni2021create}, motivating the use of heuristics and algorithms which limit annotation efforts while maximizing model learning, illustrated in Figure \ref{fig:beginning_examples}. 

\section{Related Research}

Active learning methods are driven by a query function which selects relevant data from an unlabeled pool to be annotated and joined to the training set. These methods broadly divide into two classes: uncertainty-based and diversity-based methods \cite{dasgupta2011two}. In uncertainty-based methods, data is selected by the query function's assessment of how confusing the datum is \textit{to the existing model}. On the other hand, in diversity-based methods, data is selected by being distinct from existing training data by some measure, and this can be done without consideration of the learning model. 

\subsection{The Role of Uncertainty and Diversity-Based Methods in Closed and Open Set Learning}

In closed-set learning, it is assumed that a system should classify or learn about a fixed set of target classes. By contrast, in open-set learning, the system assumes that it may encounter novel data which belongs to a class unrepresented by its current target set. Naturally, this brings up many research challenges in recognizing this novelty when it appears, determining when to define a new set construct, and integrating new constructs into the learning mechanism. 

Here, we suggest that diversity-based methods are particularly well-suited for these open-set learning tasks. Because uncertainty-based methods select relative to their existing world model, there is an inductive bias imposed in relating new data to existing patterns. On the other hand, in diversity-based methods, data is compared only to other data, analogous to unsupervised learning. This does create a tradeoff: closed-set learning excels under uncertainty-driven sampling, since these methods are optimized for the current world model and target set, but cannot treat the world as ``open" as diversity-driven sampling. But, critically, we show in this research that diversity-based active learning still provides a strong benefit to the learning system (even if not ``optimal" to the particular model and set definition), \textit{and} is suitable for open-set data selection. 

\begin{figure*}
    \centering
    \includegraphics[width=.99\textwidth]{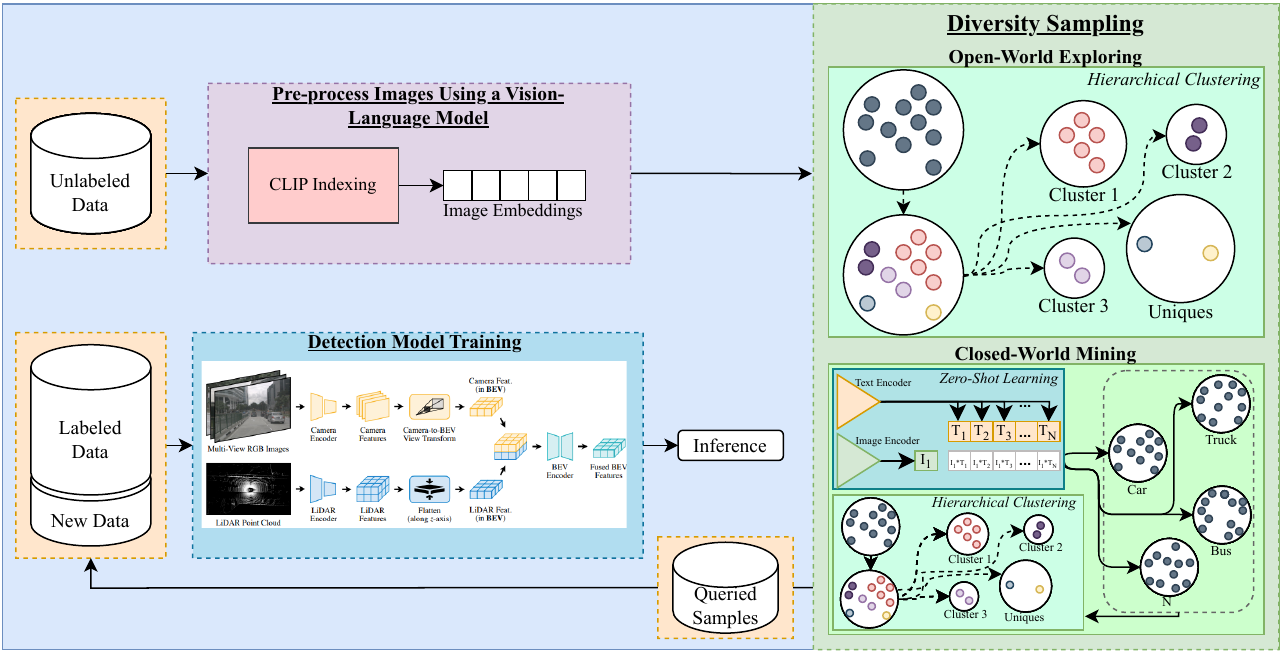}
    \caption{VisLED System Overview. For both Open-World Exploring and Closed-World Mining, the system begins with the processing of the unlabeled data pool into vision-language embedding representations. In Open-World Exploring, these embeddings are clustered and used as the basis for a query. In Closed-World Mining, the embeddings are first used in zero-shot learning to classify scenes based on object appearance, and then further clustered per-class, offering a chance to sample from particular classes which are known to be minority in the labeled training set.}
    \label{fig:architecture}
\end{figure*}

\subsection{Learning from Vision-Language Representations}

Prior research has shown that vision-language representations such as embeddings from contrastive language-image pretraining (CLIP) \cite{radford2021learning} can be used to identify novelty of an image relative to a set (and, as a bonus, can be decoded into a verbal explanation of novelty) \cite{greer2024towards}. In our research, we utilize this representation and corresponding ability to select novel images as a proxy for the amount of useful, previously-unexplored information within a complete multimodal driving scene, allowing for an active learning query to select diverse samples based on vision-language encodings of scene images.

\section{Algorithm}

Here, we present our algorithm named Vision-Language Embedding Diversity Querying (VisLED-Querying), which can be viewed in Figure \ref{fig:architecture}. The algorithm can be used in two different settings:
\begin{enumerate}
    \item Open-World Exploring: this method imposes no particular class expectations on the data. It is suitable for cases when the model seeks to include information which is most novel relative to data it has seen previously. 
    \item Closed-World Mining: this method utilizes a zero-shot learning \cite{radford2021learning} step to sort data between a fixed set of classes before evaluating for novelty, filtering any points estimated to not belong to one of the closed-set classes. This method is suitable for mining new and different instances of existing classes, but may also filter out the most difficult or unusual instances even from known classes if the zero-shot method fails to recognize the object. 
\end{enumerate}

\begin{algorithm}
\SetAlgoLined
\caption{Open-World Exploring VisLED-Querying}
\KwIn{Unlabeled pool of egocentric driving scene images}
\KwOut{Updated training set}
Embed each egocentric driving scene image from the unlabeled pool using CLIP\;
Use hierarchical clustering to separate the embeddings\;
Sample new data points from the unclustered set for addition to the training set\;
\end{algorithm}

\begin{algorithm}
\SetAlgoLined
\caption{Closed-World Mining VisLED-Querying}
\KwIn{Unlabeled pool of egocentric driving scene images}
\KwOut{Updated training set}
Embed each egocentric driving scene image from the unlabeled pool using CLIP\;
Encode each class label using a text encoding\;
Applying zero-shot learning by maximizing the product of the embeddings, sort the embedded images by class\;
For each class, apply hierarchical clustering\;
Sample new data points from the unclustered set associated with the desired class, and add to the training set\;
\end{algorithm}

In the closed-world mining setting, when employing CLIP's \cite{clip} zero-shot learning technique for classification, the algorithm examines each sample image to identify objects which are predicted to belong to one or more of the model's predefined classes. Each sample is assigned to a single class, in this case taken as the argmax class over all classes considered using the zero-shot learning method. We note that, in our experiments, this method predominantly identifies one class with high accuracy. In instances where other classes may also be identified, their confidence scores are typically low enough to risk false positives, rendering them inadequate for threshold-based classification; therefore, we use a single-class assignment for simplicity and accuracy. We do note that, as an algorithm variant, it is reasonable to distribute scene images to multiple classes if respective confidence values for the additional classes are sufficiently high. 

Once the samples for each class have been identified, embeddings will be generated separately for each class, followed by hierarchical clustering. Subsequently, a number of samples will be selected from each class, with a focus on sampling from clusters with minimal data representation. Initially, the algorithm will prioritize unique samples (clusters with only one sample present), matching them with corresponding scene names until the desired number of unique scenes is achieved in the training set. Upon inclusion of all scene-names from unique samples, the algorithm will proceed to clusters containing pairs of images, and so on, until the required number of scenes have been sampled for the training set.

In the open-world exploring setting, this same procedure is followed beginning with sampling embeddings from unique singleton clusters, without any pre-classification step to prioritize drawing from particular classes. 

\begin{figure}
    \centering
    \includegraphics[trim={0 0 36.5cm 0}, clip,width=.48\textwidth]{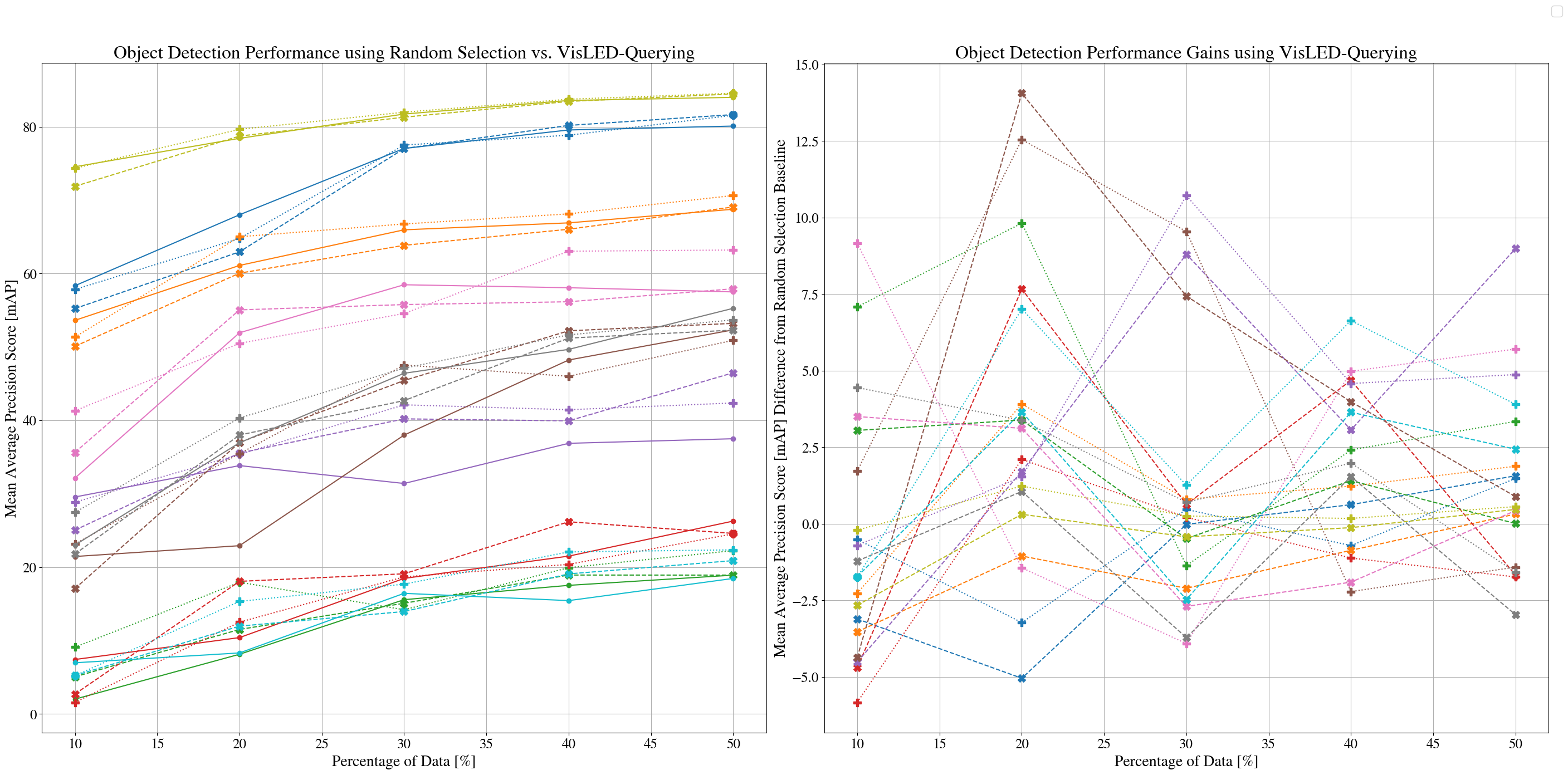}
    \includegraphics[trim={35cm 0 .5cm 0}, clip,width=.48\textwidth]{figures/combined_graphs.png}
    \includegraphics[trim={29.7cm 19.5cm 0 0}, clip, width = .17\textwidth]{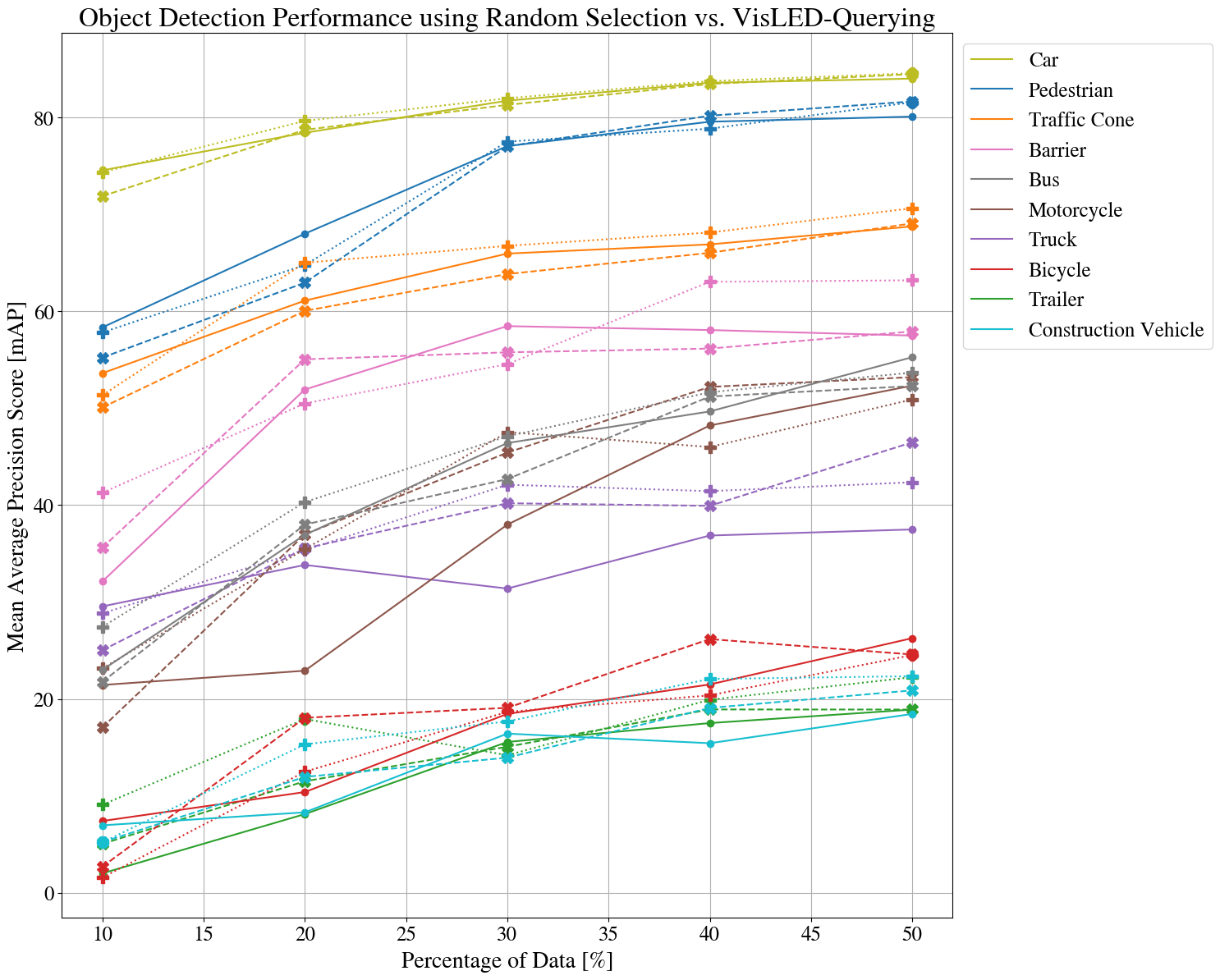}
    \caption{BEVFusion models are trained using three different data selections: Random (dot markers and solid line), VisLED-Closed-World (x-markers and dashed line), and VisLED-Open-World (+-markers and dotted line). The top graph illustrates detection performance, while the bottom graph illustrates performance difference relative to the random-selection baseline. Performance is averaged over 5 complete data selection + training runs of each model at each training pool size.}
\label{fig:overall_performance}
\end{figure}

\section{Experimental Evaluation}

\subsection{Dataset}

We use the nuScenes object detection dataset \cite{caesar2020nuscenes} for our experiments. nuScenes contains 1.4M camera images and 400k LIDAR sweeps of driving data, originally labeled by expert annotators from an annotation partner. 1.4M objects are labeled with a 3D bounding box, semantic category (among 23 classes), and additional attributes. nuScenes comprises 1000 scenes. In order to maintain complete control over the scenes within the dataset, we modify the fundamental database setup slightly, using the method introduced in \cite{ghita2024activeanno3d, greer2024and} to accommodate active learning queries. We use the \textit{trainval} split of the dataset for public reproducibility. 

\subsection{3D Object Detection Model}
We explore the BEVFusion approach to 3D object detection \cite{liu2023bevfusion}, which has demonstrated notable performance, ranking third in the nuScenes tracking challenge and seventh in the detection challenge, and the top performing method which has been made publicly reproducible. While various methods exist to integrate image and LiDAR data into a unified representation, LiDAR-to-Camera projection methods often introduce geometric distortions, and Camera-to-LiDAR projections face challenges in semantic-orientation tasks. BEVFusion addresses these issues by creating a unified representation that preserves both geometric structure and semantic density.

In our implementation, we utilize the Swin-Transformer \cite{swinT} as the image backbone and VoxelNet \cite{VoxelNet} as the LiDAR backbone. To generate bird's-eye-view (BEV) features for images, we employ a Feature Pyramid Network (FPN) \cite{FPN} to fuse multi-scale camera features, resulting in a feature map one-eighth of the original size. Subsequently, images are down-sampled to 256x704 pixels, and LiDAR point clouds are voxelized to 0.075 meters to obtain the BEV features necessary for object detection. These modalities are integrated using a convolution-based BEV encoder to mitigate local misalignment between LiDAR-BEV and camera-BEV features, particularly in scenarios of depth estimation uncertainty from the camera mode. We provide a comprehensive overview of the architecture, including its integration with VisLED-Querying, in Figure \ref{fig:architecture}.

\subsection{Experiments and Results}

We train the BEVFusion model in increasing training set sizes of 10\% increments, using four different acquisition modes: (1) Random Sampling, (2) Entropy-Querying, (3) VisLED-Querying with Closed-Set Mining setting, and (4) VisLED-Querying with Open-World Exploring setting. We repeat each VisLED method four times at each data pool size, taking the average performance from four trials. 

As hypothesized, active learning strategies outperform the random baseline, and the entropy-querying method is dominant due to its nature of optimizing uncertainty with respect to the model, as opposed to VisLED's model-agnostic sampling. Yet, as illustrated in Table \ref{tab:nuscenesprogress_random_entropy}, VisLED still stays consistently ahead of random sampling, and offers a 1\% gain over random sampling mAP at 50\% of the data pool, \textit{all without requiring \textbf{any} model training or inference}. Interestingly, the open-world exploration setting tends to marginally outperform the closed-world mining setting at nearly all data pool sizes for both metrics, suggesting that the novelty represented in the language embeddings is sufficient for identification of informative samples, even without inducing any bias from categorizing samples beforehand. On the other hand, it is also possible that the uncertainty in classifying the objects being mined for in fact makes these objects \textit{less} likely to be found in the closed-world mining setting, again encouraging the use of the open-world exploring setting in any case.  

\begin{figure}
    \centering
    \includegraphics[trim={0 0 5.5cm 0}, clip,width=.5\textwidth]{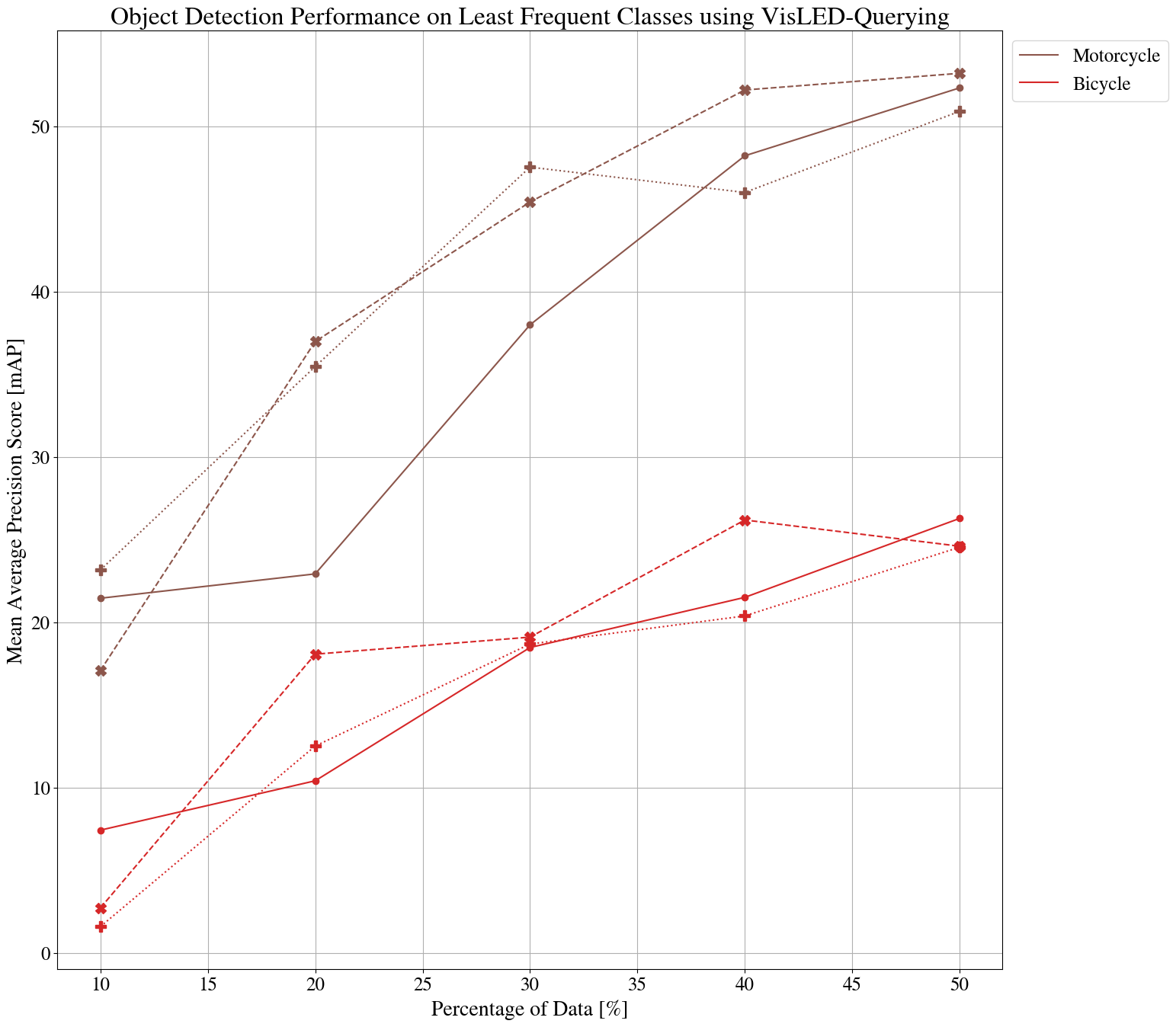}
    \caption{The bicycle and motorcycle classes are least represented in the nuScenes dataset, which causes these classes to appear infrequently during training when selecting data with random sampling. By using VisLED to sample, more bicycle and motorcycle instances are drawn, leading to a performance gain at early data increments. This gain levels off as the training pool aggregates all bicycle and motorcycle samples.}
    \label{fig:least}
\end{figure}

\begin{figure}
    \centering
    \includegraphics[trim={0 0 4cm 0}, clip,width=.5\textwidth]{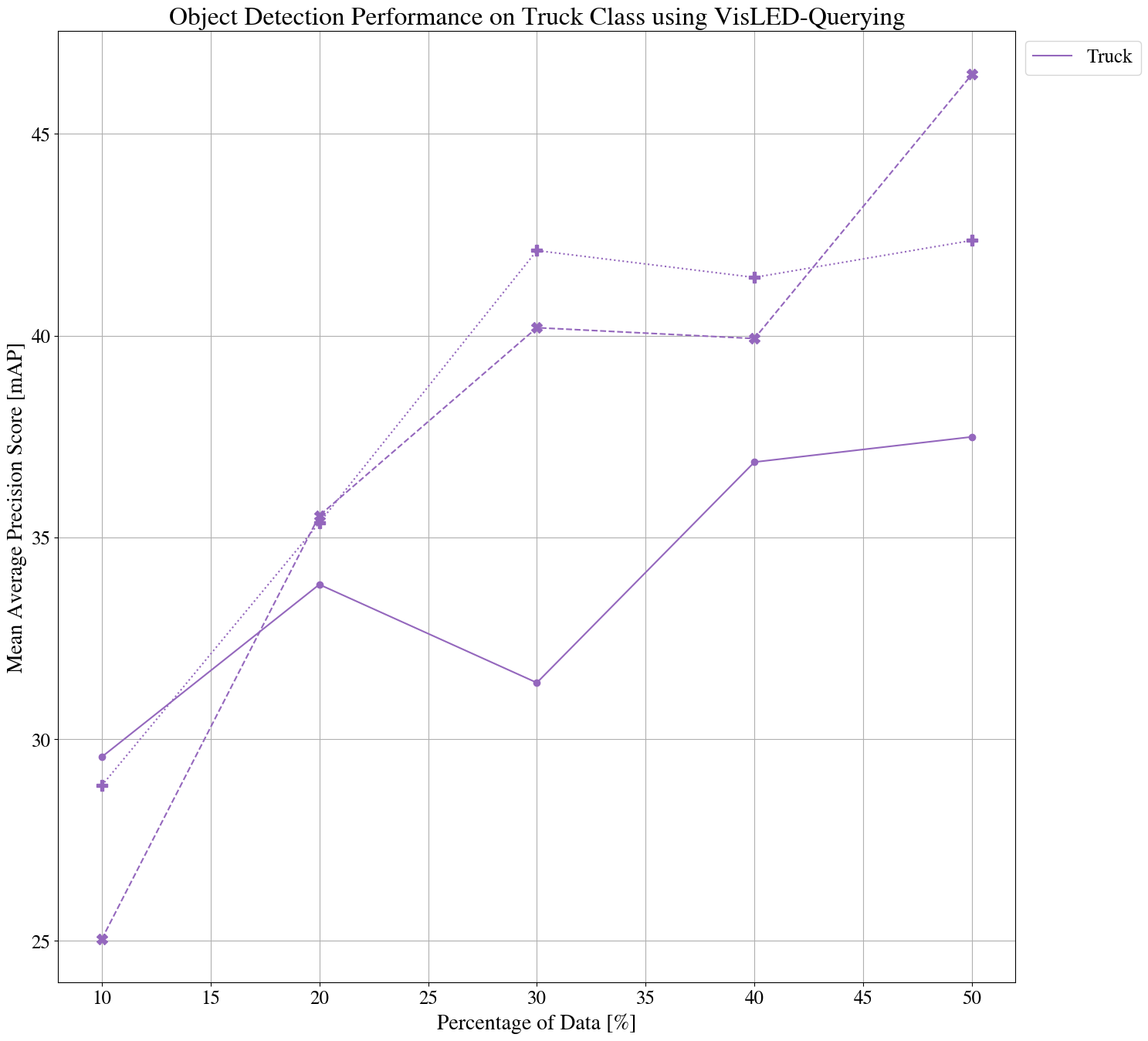}
    \caption{Detection performance on the truck class provides a clear illustration of the benefits of VisLED querying. Of special interest is the fact that the truck class would be nearly completely sampled around 70\% nuScenes training pool size, using the uniform sampling scheme of closed-world-VisLED; in other words, once all instances of a particular class are sampled, the benefit will begin to level off.}
    \label{fig:truck}
\end{figure}

Per-class performance is illustrated in Figure \ref{fig:overall_performance}. As expected, class performance correlates with class representation in the nuScenes dataset. Observing the differences in VisLED-selected detection performance over the random selection baseline in the bottom graph of Figure \ref{fig:overall_performance} reveals some interesting patterns; at 10\% data, 4 of the 10 classes perform above random. At 20\% data, this increases to 7 of 10 classes above random, and by significantly higher margins of benefit than the opposing margins of detriment when underperforming. The same ``benefits-outweigh-costs" pattern repeats at the other data levels. The particular spike in performance around 20\% data may also have an interesting explanation, which relates to performance on the two least-represented classes, illustrated in Figure \ref{fig:least}. These classes, motorcycle and bicycle, represent 1.08 and 1.02\% of the nuScenes objects, respectively. When VisLED-CW is used to sample uniformly from each class, it would actually \textit{run out} of motorcycle and bicycle samples around 20\% training data, because at each 10\% data increment, 1\% of nuScenes data should be coming from each of the 10 classes. This means that after two training rounds, the data from the particular class should be exhausted, which explains why we see the greatest margin in performance over random happening at this level - and, this is a strong gain, around 10\% mAP for bicycle and 5\% mAP for motorcycle. This further explains the asymptotic behavior we see as the data volume approaches 50\%; there is less prototypical data for these classes available for the detector to learn at this point. For similar reason, we see a consistent boost in the performance on the truck class (illustrated in Figure \ref{fig:truck}); this class has 7.59\% representation in nuScenes, making it an excellent candidate for uniform gain throughout all training rounds, and reaching almost its entire representative dataset by the 50\% data round. This balance of data proportionality and sampling may explain the consistent gains, even as high as 20\% mAP improvement over baseline at intermediate rounds. 

\begin{figure}
    \centering
    \includegraphics[trim={0 0 8cm 0}, clip,width=.5\textwidth]{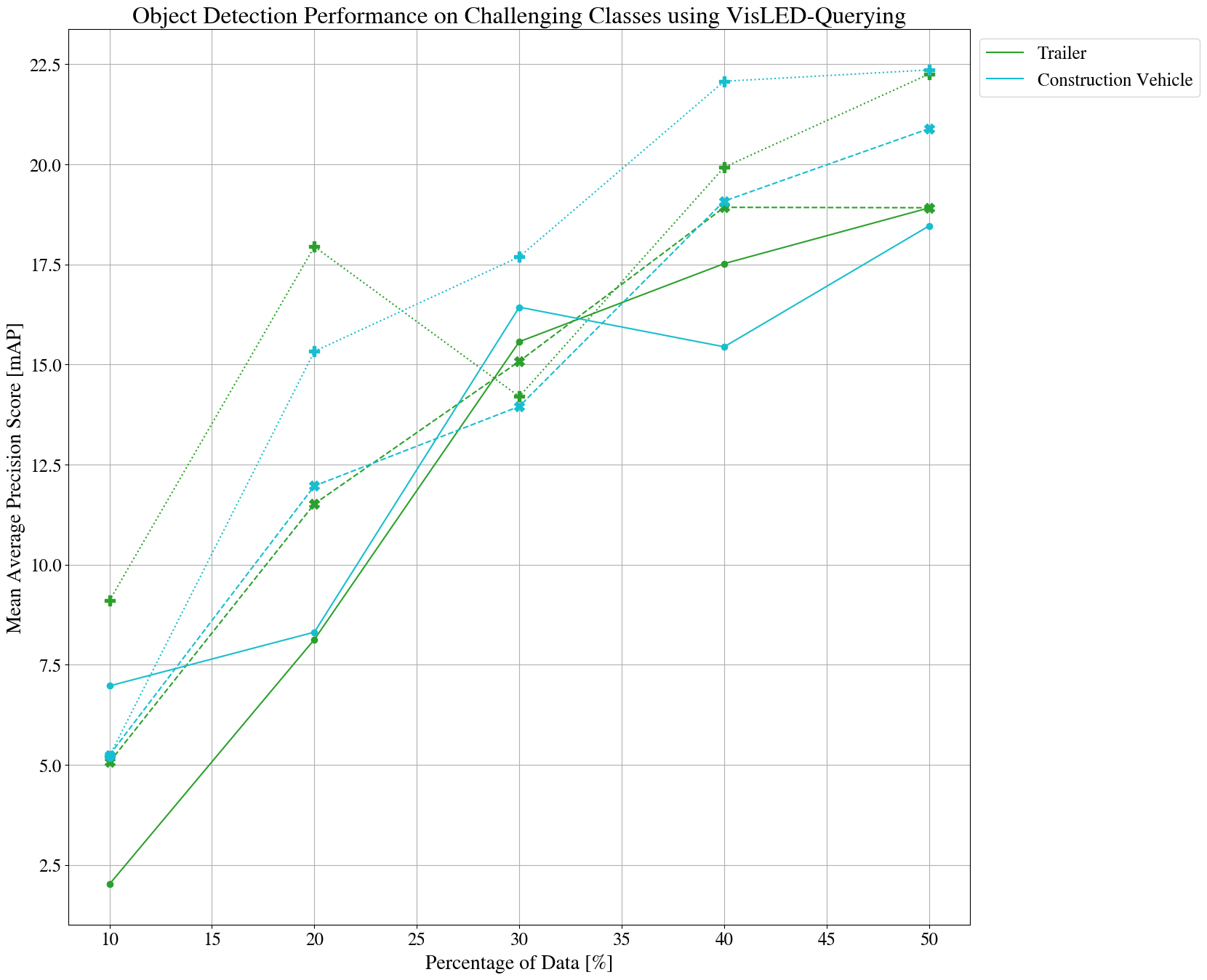}
    \caption{From class performance, the trailer and construction vehicle classes are most challenging to learn. When VisLED querying is used, informative samples from these classes are pulled into the training pool, giving stronger detection performance than random sampling at nearly all data volumes.}
    \label{fig:hardest}
\end{figure}

Besides the issue of dataset representation, we can also examine performance on classes which may be generally difficult to learn. Looking at the two lowest-performing classes on baseline (trailer and construction vehicle, representing 2.13 and 1.26\% of nuScenes respectively), Figure \ref{fig:hardest} shows that these classes indeed benefit from VisLED sampling - in fact, at 20, 40, and 50\% training data, both closed-world and open-world methods dominate the random sampling selection method. 

\begin{table*}[]
    \centering
    \caption{This table shows the mean average precision (mAP) and nuScenes detection score (NDS) metrics for the random sampling, and VisLED-querying (Closed-World Mining and Open-World Exploring) in every round. It also shows the mAP and NDS scores for the full training split when trained using one GPU.}
    \begin{adjustbox}{width=\textwidth}
    \begin{tabular}{c|c|c|c|c|c|c|c|c|c|c|c}
        % \hline
        % \multicolumn{2}{c|}{Labeled Pool} & \multicolumn{2}{|c|}{Random} & \multicolumn{2}{|c}{Entropy} & \multicolumn{2}{|c}{VisLED}\\
        \hline 
        \multicolumn{2}{c|}{Labeled Pool} & \multicolumn{5}{|c|}{mAP} & \multicolumn{5}{|c}{NDS} \\
        \hline
        Rounds & \% & Random & \multicolumn{2}{|c|}{VisLED (CWM)} & \multicolumn{2}{|c|}{VisLED (OWE)} & Random & \multicolumn{2}{|c|}{VisLED (CWM)} & \multicolumn{2}{|c}{VisLED (OWE)} \\
        \hline
        \multicolumn{2}{c|}{} & & Mean & STD & Mean & STD & & Mean & STD & Mean & STD \\
        % Rounds & \% & mAP $\uparrow$ & NDS $\uparrow$ & mAP $\uparrow$ & NDS $\uparrow$ & mAP $\uparrow$ & NDS $\uparrow$ \\
        \hline \hline 
        % \rule{0pt}{2ex} 1 & 2.52\% \\
        % \hline
        \rule{0pt}{2ex} 1  & 10\% & 30.95 & 28.94 & 0.37 & \textbf{32.14} & 0.76 & 33.53 & 32.59 & 0.33 & \textbf{34.85} & 0.71 \\
        %\rule{0pt}{2ex} 1  & 10\% & 0.3095 & 0.3353 & 0.3106 & 0.3409 & 29.14 & 32.16 \\
        \hline
        % \rule{0pt}{2ex} 6 & 15\% & 0.3419 & 0.3679 & 0.3639 & 0.3868 \\
        % \hline
        \rule{0pt}{2ex} 2 & 20\% & 38.00 & 40.61 & 0.94 & \textbf{41.70} & 0.95 & 40.14 & 41.34 & 0.56 & \textbf{42.44} & 0.96 \\
        %\rule{0pt}{2ex} 2 & 20\% & 0.38 & 0.4014 & 0.4041 & 0.4185 & 40.76 & 41.18 \\
        \hline
        % \rule{0pt}{2ex} 10 & 25\% & 0.4236 & 0.4497 & 0.4217 & 0.4446 \\
        % \hline
         \rule{0pt}{2ex} 3 & 30\% & 44.94 & 45.28 & 0.93 & \textbf{46.94} & 0.25 & 48.41 & 48.82 & 0.86 & \textbf{50.84} & 1.16 \\
        %\rule{0pt}{2ex} 3 & 30\% & 0.4494 & 0.4841 & 0.4557 & 0.5011 & 45.01 & 49.40 \\
        \hline
        % \rule{0pt}{2ex} 14 & 35\% & 0.4474 & 0.4736 & \textbf{0.4676} & \textbf{0.5181} \\
        %\hline
        \rule{0pt}{2ex} 4 & 40\% & 47.73 & 49.26 & 0.53 & \textbf{49.59} & 0.66 & 53.10 & \textbf{53.64} & 0.32 & 52.99 & 0.59 \\
        %\rule{0pt}{2ex} 4 & 40\% & 47.73 & 53.1 & 0.4924 & 0.5380 & 49.21 & 53.64 \\
        \hline
        \rule{0pt}{2ex} 5 & 50\% & 49.90 & 50.98 & 0.13 & \textbf{51.74} & 1.08 & 55.64 & 56.40 & 0.40 & \textbf{56.61} & 1.09 \\
        %\rule{0pt}{2ex} 5 & 50\% & \textbf{49.9} & \textbf{55.64} & \textbf{63.88} & \textbf{64.85} & \textbf{51.05} & \textbf{56.45} \\
        \hline
        \hline
         & 100\% &\multicolumn{5}{|c|}{52.88} & \multicolumn{5}{|c}{58.73}\\
        \hline
    \end{tabular}
    \end{adjustbox}
%Both VisLED methods outperform random sampling consistently, and reach nearly the same level of performance as 100\% of the data at just the 50\% data point, showing faster learning than the baseline method.}
    \label{tab:nuscenesprogress_random_entropy}
\end{table*}

\section{Discussion and Conclusion}

Our presented learning method, VisLED-Querying, samples without any information about the model. This enables VisLED to select novel, informative data points, to the extent that novelty is visibly identifiable, for \textit{any} model. The benefit this offers is that a data point may need to be annotated only once, and can then be used in a variety of models for additional autonomous driving tasks instead of sampling and possibly forming an entirely different set for annotation. While these gains may be marginal in the current data setting ($<$ 1000 scenes), at scale, these performance gains may translate to serious reductions in annotation costs and safety-critical detection failures. Further, VisLED offers one key possibility that is otherwise limited on uncertainty-driven approaches: VisLED will recommend unique samples without any prior assumptions on class taxonomy, making it especially suited to open-set learning, where new classes may be introduced at any time. This capability, when paired with methods of self- or semi-supervised learning for object detection by fusing LiDAR and camera \cite{hekimoglu2024monocular}, may prove especially beneficial in identifying and learning from novel encounters. In future research, we plan to experiment on the effectiveness of VisLED in multi-task learning settings \cite{hekimoglu2023multi}, experiments on other benchmark datasets \cite{zimmer2023tumtraf}, and experiments in open-set and continual learning. Further, experiments will also examine the benefits of VisLED querying over safety-critical underrepresented classes in driving scenes, such as pedestrians using a stroller (0.09\% of nuScenes objects), mobility aid (0.03\%), or wheelchair (0.04\%), or emergency vehicles such as an ambulance (0.00004\%). In these cases, the ability to use zero-shot learning methods or even the general, open-world VisLED querying approach, may lead to training data which is more balanced and effective at capturing data which sits on the long tail of driving scenarios, making for safer perception and planning.

\bibliographystyle{unsrt}
\bibliography{bare_jrnl}

\end{document}